\documentclass[letterpaper, 10 pt, conference]{ieeeconf}  
\usepackage{xcolor}
\usepackage{wrapfig}
\usepackage{graphicx}
\usepackage{biblatex} 
\usepackage{float}
\usepackage{caption}
\usepackage{subcaption}
\usepackage{amsmath}
\usepackage{amssymb}
\IEEEoverridecommandlockouts                              

\title{\LARGE \bf
A Semi-supervised Object Detection Algorithm for Underwater Imagery
}

\author{Suraj Bijjahalli$^{1}$,Oscar Pizarro$^{1,2}$, and Stefan B. Williams$^{1}$%
\thanks{$^{1}$The authors are with the Australian Centre for Field Robotics, University of Sydney, NSW Australia
        {\tt\small (s.bijjahalli, o.pizarro, stefanw)@acfr.usyd.edu.au}}%
\thanks{$^{2}$O. Pizarro is also with the Marine Technology Department, Norwegian University of Science and Technology, Trondheim, Norway}%
\thanks{*The research for this paper received funding from the Australian Government through Trusted Autonomous Systems, a Defence Cooperative Research Centre funded through the Next Generation Technologies Fund.The authors also thank Thales Australia for supporting this work.}
}
\begin{document}

\maketitle
\thispagestyle{empty}
\pagestyle{empty}

\begin{abstract}

Detection of artificial objects from underwater imagery gathered by Autonomous Underwater Vehicles (AUVs) is a key requirement for many subsea applications. Real-world AUV image datasets tend to be very large and unlabelled. Furthermore, such datasets are typically imbalanced,  containing few instances of objects of interest, particularly when searching for unusual objects in a scene. It is therefore, difficult to fit models capable of reliably detecting these objects. Given these factors, we propose to treat artificial objects as anomalies and detect them through a semi-supervised framework based on Variational Autoencoders (VAEs).  We develop a method which clusters image data in a learned low-dimensional latent space and extracts  images that are likely to contain anomalous features. We also devise an anomaly score based on extracting poorly reconstructed regions of an image. We demonstrate that by applying both methods on large image datasets, human operators can be shown candidate anomalous samples with a low false positive rate to identify objects of interest. We apply our approach to real seafloor imagery gathered by an AUV and evaluate its sensitivity to the dimensionality of the latent representation used by the VAE. We evaluate the precision-recall tradeoff and demonstrate that by choosing an appropriate latent dimensionality and threshold, we are able to achieve an average precision of 0.64 on unlabelled datasets.
\end{abstract}

\section{INTRODUCTION}

Detection and localization of artificial underwater objects in optical imagery captured by Autonomous Underwater Vehicles (AUVs) is a key requirement for many industrial, scientific and defence applications. Missions conducted by AUVs generate a large volume of imagery, which makes it impractical for human operators to visually inspect and isolate a sparse set of artificial objects during the post-mission phase. 
There is therefore a need for a robust and efficient automated vision-based method for detecting artificial objects in a variety of benthic habitats. Most underwater object detection algorithms are largely derived from traditional pattern recognition methods. In most of these methods, the properties of the object are known beforehand, and it is assumed that datasets with ground-truth labels can be sourced to train a classifier. With the advent of deep learning, features can be learned rather than crafted by hand. However, most deep learning algorithms rely on the availability of large, well-balanced and labelled datasets. This type of supervised pipeline presents the following problems: (1) hand-labelling of large-scale image datasets is time-consuming, cumbersome and prone to error; (2) Underwater datasets with artificial objects tend to be highly imbalanced i.e., a small fraction of images contain artificial objects, with the bulk of the images corresponding to naturally occurring features. If the primary aim is not to explicitly predict a single, specific class out of several possible classes, but more broadly to separate artificial objects from naturally occurring features, the characteristics of this problem are better suited to an unsupervised or semi-supervised anomaly detection framework. In this paper, we focus on using autoencoders to detect a minority class (anomalous images containing artificial objects) from a large unlabelled database comprising mostly the majority class (natural images). Autoencoders are essentially unsupervised generative models. Once trained, the encoder component of an autoencoder can be used as an unsupervised feature extractor for seafloor imagery \cite{rao2017multimodal,flaspohler2017feature}, which can outperform handcrafted features \cite{yamada2021learning}. Autoencoders have previously been used for underwater anomaly detection as well \cite{zurowietz2018maia}. However, most approaches focus on the use of the L2 loss (reconstruction error) as a detection metric - The underlying assumption is that anomalous data which is sparsely observed or completely absent from the training dataset cannot be reconstructed accurately when observed in the field.
In our proposed approach, we choose to avoid relying solely on reconstruction loss. We additionally make use of the latent space of the autoencoder in which anomalous images are separable through a clustering-based method.  We base our approach on Variational Autoencoders (VAE), a probabilistic extension of autoencoders with greater flexibility and control over the distribution of the latent space. 
We make the following primary contributions:
\begin{itemize}
\item Development of a VAE model that can reconstruct seafloor imagery;
\item Investigation of a clustering method to reduce large datasets to a smaller subspace for aiding anomaly detection;
\item  Investigation of L2 loss as an anomaly detection metric, and the development of a novel Region Of Interest (ROI) detection metric;
\item Analysis of detection/false alarm rate sensitivity to latent vector dimensionality; 
\end{itemize}

\begin{figure*}[h!]
  \centering
  \includegraphics[width=0.70\textwidth]{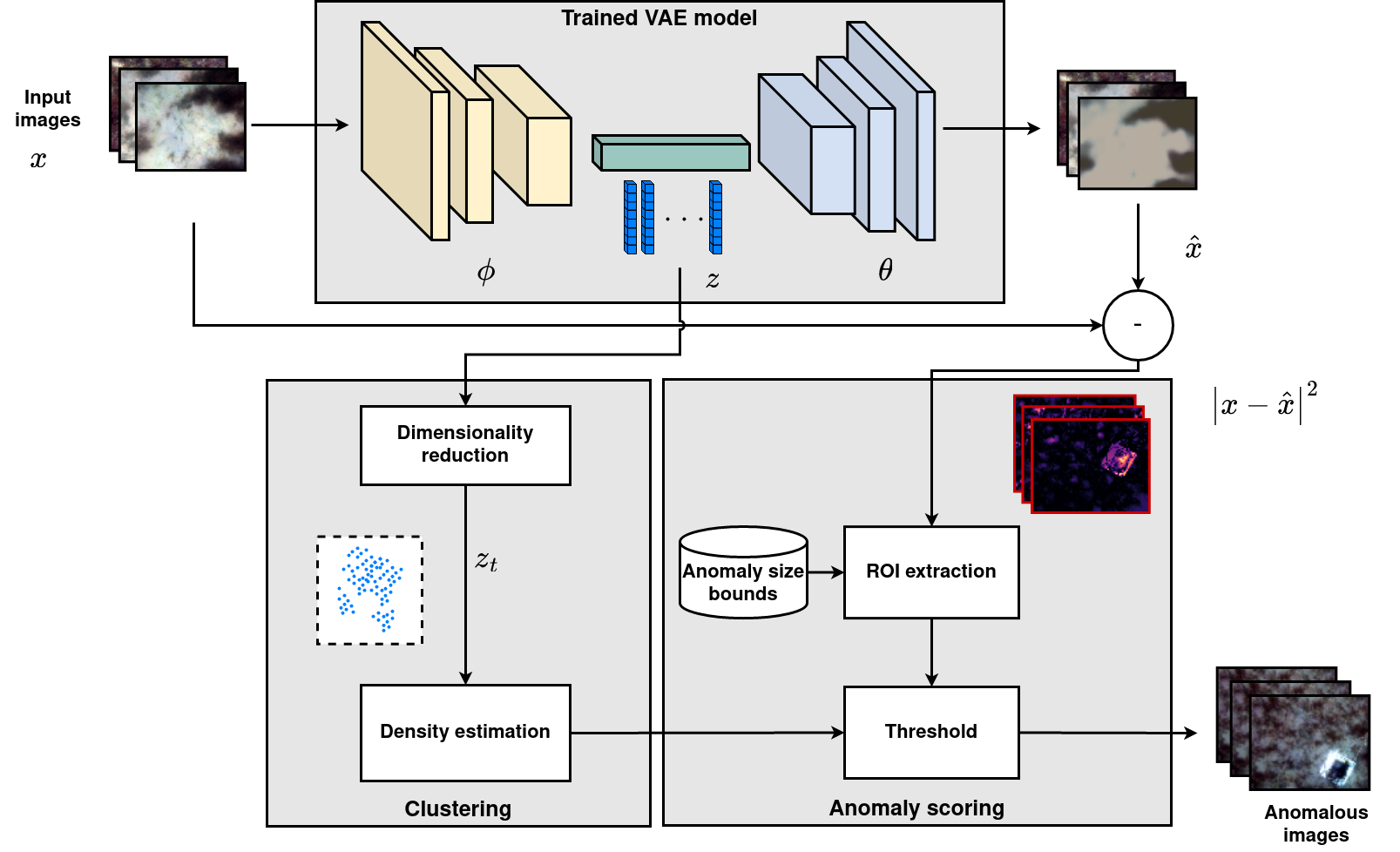}
  \caption{System overview - Images captured by an AUV are passed as inputs to a trained VAE. Low density regions of the latent space are extracted and further queried using an anomaly map obtained from the VAE reconstruction error.} \label{fig: system overview}
\end{figure*}

The remainder of this paper is structured as follows: Section \ref{related work} summarizes related work and highlights some of the key research gaps in this domain; Section \ref{method} presents an overview of our proposed method, followed by a brief description of the foundation of VAEs in Section \ref{VAE math}. We briefly describe the datasets we use and our training pipelines in  Section \ref{datasets}. In Section \ref{detection metrics}, we show the shortcomings behind using L2 reconstruction loss as an anomaly detection metric before presenting our proposed metrics.
We demonstrate the performance of our proposed method in Section \ref{results}, along with a sensitivity analysis of performance against model capacity.

\section{Related work}\label{related work}

A broad survey of deep learning-based underwater object detection algorithms can be found in \cite{teng2020underwater}. In \cite{yu2018man}, a transfer learning approach is employed, where a pre-trained object detection model is repurposed by means of a synthetically generated dataset. Outside the underwater domain, anomaly detection in images using autoencoders is well-researched, and several approaches can be adopted to our application. Each image is essentially assigned an anomaly score, which is thresholded to flag anomalies. Most anomaly scores are typically based on reconstruction error \cite{fei2020attribute, bozcan2021gridnet}. More specifically, the L2 loss between the input image and the model reconstruction is used as an anomaly score. The underlying assumption is that the L2 loss is small for in-distribution images and large for out-of-distribution images. As we show in \ref{detection metrics}, this assumption rarely holds true for images gathered in the field in practice. Depending on the choice of model architecture and capacity, it is possible for the model to reconstruct the anomaly even though it has only been trained on in-distribution samples \cite{havtorn2021hierarchical}. To address this problem, in \cite{huang2021self}, a transformation is first applied to the image, and then the model is trained to reconstruct the untransformed image. This forces the model to learn the high-level visual semantic features of in-distribution images. Apart from reconstruction error, anomaly scores can be obtained based on restoration error \cite{chen2020unsupervised} or latent space distances \cite{zong2018deep, mantegazza2022outlier, abati2019latent}.  

\section{Method} \label{method}
We design our methodology to flag and extract anomalous samples from AUV-gathered imagery datasets in a semi-supervised fashion. To make our terminology consistent within this paper, we use the terms outlier and anomalous images interchangeably to refer to images of the seafloor with artificial objects in them. 
Our method makes the following assumptions: (1) An existing database of inlier images is available to train a VAE. This assumption is well justified, since most imagery from AUV missions contains a large number of inlier images with few (if any) anomalies; (2) The size of the anomalous object to be detected is approximately (upper and lower size bounds) known;(3) The intrinsic  and extrinsic parameters of the camera are known. This is so that we can estimate the angular Field of View(FOV) of the camera and translate the object size bounds (in squared metres) to pixel area bounds on the images;(4) We also assume that anomalous samples are likely to gather in the low density regions of the VAE latent space. We justify this assumption in sections \ref{latent space clustering} and \ref{results}, where we demonstrate this experimentally. The high-level architecture of our system is illustrated in  Fig. \ref{fig: system overview}. Our proposed method is centred on the use of a VAE trained entirely on inlier images i.e., images that do not contain an anomaly. We forward-propagate each image gathered from the AUV mission through our trained VAE. The encoder extracts features from the image and downsamples it through a series of convolutions which compress the input image to a low-dimensional latent vector $z$. The decoder is a generative model that attempts to reconstruct the input image from $z$, which in turn is the output of a sampling process. The encoder essentially learns a distribution (assumed to be Gaussian in our case). The feature maps in the final layer of the encoder are flattened into two vectors ($d$x1 dimensions) corresponding to a mean ($\mu_z$) and standard deviation ($\sigma_z$). The latent vector is then sampled from this multi-dimensional Gaussian, and serves as a $d$-dimensional representation of the input image. The decoder receives as input the latent embedding of the input image and then reconstructs it through a series of transpose convolutional operations. The pixel-wise difference between the reconstructed image and the input image is computed to generate an anomaly heatmap. We extract regions of interest (ROI) from the anomaly heatmap. Given estimates of the camera FOV and the altitude from the seafloor, we then extract only the ROIs that conform to the expected size bounds of the anomaly. Moments of the extracted ROIs are then used as an additional feature to separate inlier images from anomalous ones. In parallel, we apply Kernel Density Esimation (KDE) to estimate the density of the latent space (see Sec. \ref{latent space clustering}). KDE performance is known to worsen exponentially with high dimensional data sets \cite{scott1991feasibility}. To overcome this problem, we further reduce the dimensionality of the latent vectors. The low-density subspace of this reduced feature set is then extracted. As we demonstrate in sections \ref{latent space clustering} and \ref{results}, this subspace has a high probability of containing the anomalous images within the dataset. To reduce the number of false-positives within this extracted subspace, we further apply a threshold to the statistical moments of the ROIs obtained from the anomaly heatmaps in the previous step. We show in Sec. \ref{results} that this dual approach involving density estimation and reconstruction error greatly reduces the number of false positives.

\section{Variational Autoencoder} \label{VAE math}
VAEs are essentially generative models based on a probabilistic formulation of conventional autoencoders. We present a high-level overview of the mathematical foundation of VAEs, which were originally introduced in \cite{kingma2013auto}. A VAE defines a model of the form:
\begin{equation}
p_{\boldsymbol{\theta}}(\boldsymbol{z}, \boldsymbol{x})=p_{\boldsymbol{\theta}}(\boldsymbol{z}) p_{\boldsymbol{\theta}}(\boldsymbol{x} \mid \boldsymbol{z})
\end{equation}
where $\boldsymbol{z}$ is a latent representation of the input images $\boldsymbol{x}$. $p_{\boldsymbol{\theta}}(\boldsymbol{z})$ is the prior over $\boldsymbol{z}$, chosen to be Gaussian. 
$p_{\boldsymbol{\theta}}(\boldsymbol{x} \mid \boldsymbol{z})$  is a product of exponential family distributions (again chosen to be Gaussian) with parameters computed by a neural network decoder with parameters $\boldsymbol{\theta}$.
When training the model, the goal is to maximize the marginal likelihood of the data:
\begin{equation}
p_{\boldsymbol{\theta}}(\boldsymbol{x})=\int p_{\boldsymbol{\theta}}(\boldsymbol{x} \mid \boldsymbol{z}) p_{\boldsymbol{\theta}}(\boldsymbol{z}) d \boldsymbol{z}
\end{equation}
This computation is intractable, and hence the computation of the posterior is intractable as well:
\begin{equation}
p_{\boldsymbol{\theta}}(\boldsymbol{z} \mid \boldsymbol{x})=\frac{p_{\boldsymbol{\theta}}(\boldsymbol{z}, \boldsymbol{x})}{p_{\boldsymbol{\theta}}(\boldsymbol{x})}
\end{equation}
The solution is to compute an approximate posterior using a inference network i.e. the encoder with parameters $\phi$. The conventional approach is to maximize the so-called Evidence Lower Bound (ELBO), given by:
\begin{equation}
\mathrm{L}_{\boldsymbol{\theta}, \boldsymbol{\phi}}(\boldsymbol{x})=\mathbb{E}_{q_{\boldsymbol{\phi}}(\boldsymbol{z} \mid \boldsymbol{x})}\left[\log p_{\boldsymbol{\theta}}(\boldsymbol{x} \mid \boldsymbol{z})\right]-D_{\mathbb{K} \mathbb{L}}\left(q_{\boldsymbol{\phi}}(\boldsymbol{z} \mid \boldsymbol{x}) \| p_{\boldsymbol{\theta}}(\boldsymbol{z})\right)
\end{equation}

This objective function is the expected log likelihood (first term on the RHS) and a regularization term (second term on the RHS) which is essentially the Kulback-Leibler (KL) divergence. The regularization term ensures the per-sample posterior does not deviate too far from the prior in terms of KL
divergence. 
We make the following assumptions:
\begin{equation}
p(\boldsymbol{z})=\mathcal{N}(\boldsymbol{z} \mid \mathbf{0}, \mathbf{I})
\end{equation}

\begin{equation}
q(\boldsymbol{z} \mid \boldsymbol{x})=\mathcal{N}(\boldsymbol{z} \mid \boldsymbol{\mu}, \operatorname{diag}(\boldsymbol{\sigma}))
\end{equation}

This allows the KL to be computed in closed form:

\begin{equation}
D_{\mathbb{K} \mathbb{L}}(q \| p)=-\frac{1}{2} \sum_{k=1}^K\left[\log \sigma_k^2-\sigma_k^2-\mu_k^2+1\right]
\end{equation}
The log-likelihood term can be approximated by sampling:
\begin{equation}
\boldsymbol{z}^s \sim q_\phi(\boldsymbol{z} \mid \boldsymbol{x})
\end{equation}\begin{equation}
\mathrm{L}_{\boldsymbol{\theta}, \boldsymbol{\phi}}(\boldsymbol{x})=\log p_{\boldsymbol{\theta}}\left(\boldsymbol{x} \mid \boldsymbol{z}^s\right)-D_{\mathrm{KL}}\left(q_{\boldsymbol{\phi}}(\boldsymbol{z} \mid \boldsymbol{x}) \| p_{\boldsymbol{\theta}}(\boldsymbol{z})\right)
\end{equation}

The ELBO for the entire dataset is the ELBO for a single datapoint scaled by the number of samples $N$:
\begin{equation}\label{ELBO_loss}
\mathrm{L}_{\boldsymbol{\theta}, \phi}(\mathcal{D})=\frac{1}{N} \sum_{\boldsymbol{x} \in \mathcal{D}} \mathrm{L}_{\boldsymbol{\theta}, \boldsymbol{\phi}}(\boldsymbol{x})
\end{equation}
Owing to the stochastic process inherent in the VAE (the latent vector is sampled and is not deterministic), a so-called reparameterization trick is employed to backpropagate gradients during training \cite{kingma2013auto}. To assign the hyperparameters for the encoder and decoder, we make use of the findings in \cite{yamada2021learning}, and loosely follow the AlexNet architecture \cite{krizhevsky2012imagenet}. Our proposed model architecture is as follows:  The encoder applies a series of five convolutions and batch normalizations, before applying a leaky ReLU activation at each layer. The outputs of the encoder are flattened into two vectors, representing the mean and standard deviation of a Gaussian distribution, from which the latent vector is sampled. The latent space dimension $d$ is a hyperparameter that must be carefully tuned to adjust the model capacity, and ultimately the model reconstructions. We do not assign a value for $d$ upfront. Instead, we evaluate several models while varying this parameter. Qualitatively, we find that model reconstructions tend to be more blurry when reducing the latent dimensionality. The decoder is an inverse mapping of the encoder, with five transpose convolutional layers used to upsample the latent space to the original dimensions of the input image.

\section{Datasets} \label{datasets}
\begin{figure}[tb!]
     \centering
     \begin{subfigure}[b]{0.75\columnwidth}
         \centering
         \includegraphics[width=\textwidth]{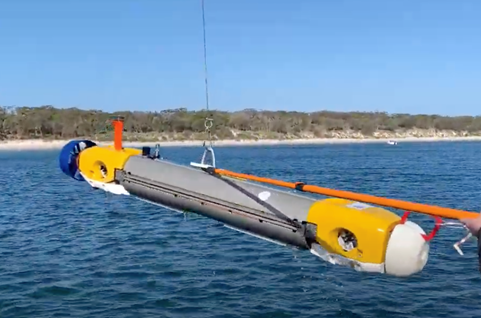}
         \caption{}
         \label{fig: nimbus}
     \end{subfigure}
     \hfill
     \begin{subfigure}[b]{0.75\columnwidth}
         \centering
         \includegraphics[width=\textwidth]{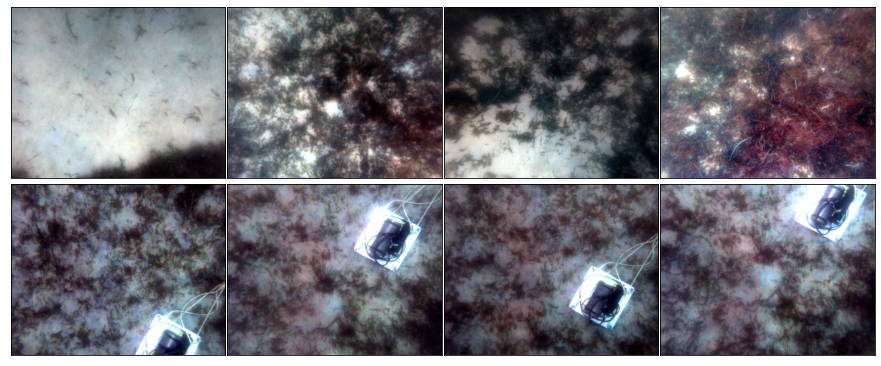}
         \caption{}
         \label{fig: sample images}
     \end{subfigure}
      \caption{ (a) \textit{Nimbus} AUV being deployed at Jervis Bay, Australia; (b) Samples of inlier images with only natural features (top row) and outlier images which contain an artificial object (bottom row). }
        \label{fig: image samples}
\end{figure}

To train the VAE, we make use of images gathered by \textit{Nimbus} (see \ref{fig: nimbus}), a hover-capable AUV developed by the University of Sydney's Australian Centre for Field Robotics (ACFR). 
We focus on imagery acquired by the AUV at Jervis Bay, Australia in April 2021. The AUV was tasked with conducting a tight lawnmower survey pattern comprised of 1m spaced track lines. At a 2m altitude, this spacing allows for full coverage of the seafloor using the vehicle’s high-resolution stereo camera (2x5Mpx) with cross-track overlap of around 25\%. The nominal 2m altitude acquires images of sufficient resolution for inspection and monitoring purposes. Following a preliminary survey of the site, an artificial anomalous `target' was introduced into the environment. This comprised a 30cm x 30cm panel with a Ultra-Short Baseline (USBL) acoustic transponder mounted on it. The survey was then repeated. The raw 16-bit imagery is colour-corrected, debayered and down-sampled to 8-bit. Only 14 images are available with the artificial target captured within the frame. These are designated as outliers, and the remaining images are designated inliers. Samples of inlier and outlier images are shown in Fig. \ref{fig: sample images}. From the preliminary dive, 14616 inlier images were acquired. We split this dataset into 70\%-30\% subsets for training and validation respectively. From a separate dive on the same campaign, we use a dataset of 614 inlier images as a test set. We then append the outlier images to the test dataset, resulting in an outlier proportion of 2.2\%.

\subsection{Model training}
To fit the model to the inlier data, we iteratively adjust the model parameters by minimizing the ELBO formulated in Equation \ref{ELBO_loss}. The images are scaled down to a 64x80 resolution, and random horizontal and vertical flip transforms are applied to augment the data. We employ an early stopping routine in which training is halted if we observe more than three epochs without a decrease in validation loss. The model is trained at a learning rate of 1e-3 on the Adam optimizer. 

\section{Anomaly detection metrics} \label{detection metrics}
Once a model is trained, we use it to generate anomaly detection metrics. As mentioned in section \ref{related work}, a commonly employed method is to evaluate the reconstruction loss. The underlying assumption is that the model will be unable to reconstruct the anomalous image well, resulting in a high reconstruction error relative to inlier images that are well represented in the training dataset. The actual anomaly score is an L2 loss to which an empirical threshold is applied. In this section, we evaluate the suitability of the L2 loss as an anomaly detection metric. We also determine whether the performance of this metric is a function of latent dimensionality. To achieve this, we train nine models by varying the latent dimension size from 8-2048 and we evaluate each model on inlier and anomalous image samples.

\subsection{L2 Reconstruction error}
\begin{figure}[h!]
     \centering
     \begin{subfigure}[b]{0.85\columnwidth}
         \centering
         \includegraphics[width=\textwidth]{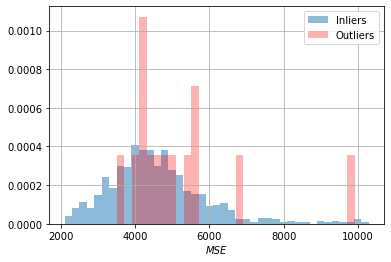}
         \caption{}
         \label{fig:MSE histogram}
     \end{subfigure}
     \hfill
     \begin{subfigure}[b]{0.95\columnwidth}
         \centering
         \includegraphics[width=\textwidth]{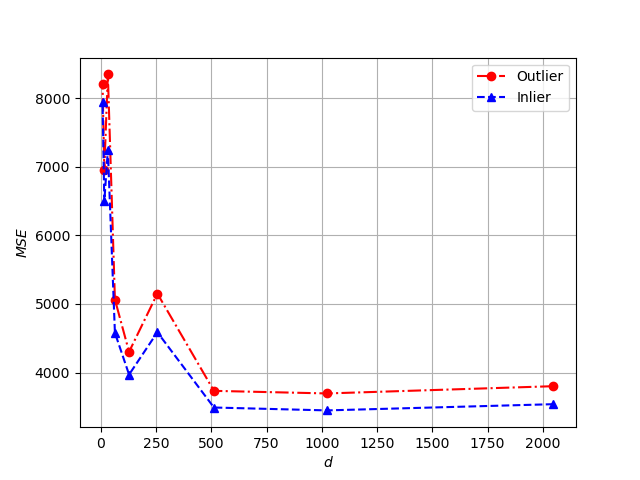}
         \caption{}
         \label{fig:MSE results}
     \end{subfigure}
      \caption{L2 anomaly scores (a) Histogram of L2 loss for model with $d$=256;(b) Mean L2 loss values for inlier and outlier images from our test dataset as a function of latent dimensionality}
        \label{fig:MSE}
\end{figure}

After training the VAE, for each image in our test dataset, we compute the L2 loss between the input image and the model reconstruction. Fig. \ref{fig:MSE histogram} shows the histogram of L2 loss for a model of latent size 256. For an input image of 64x80, this represents a compression ratio of 20:1. The distributions for the inlier and anomalous image samples are shown. It is evident that the two distributions overlap to a large degree, and that any threshold we apply would result in a large number of missed-detections or false-alarms. We find that detection performance does not significantly improve as a function of latent dimensionality. Fig. \ref{fig:MSE results} shows the mean of the anomaly scores obtained by varying the size of the bottleneck. We observe that there is no appreciable increase in separation between the L2 loss for inliers and outliers regardless of the latent dimensionality. 
From this, we conclude that the approach of using an L2 loss as an anomaly score yields poor results for complex seafloor datasets in which there is a highly diverse set of habitats i.e. sand, kelp, and rocks. We find that when the object to be detected (the anomaly) occupies a small portion of the image, the reconstruction loss of the surrounding patch tends to blunt the anomaly score of the overall image. 
On the other hand, if we visually inspect the squared-difference between the images and their reconstructions, the outlier images are readily distinguishable from the inliers. 
\begin{figure}[htb!]
     \centering
     \begin{subfigure}[b]{0.85\columnwidth}
         \centering
         \includegraphics[width=\textwidth]{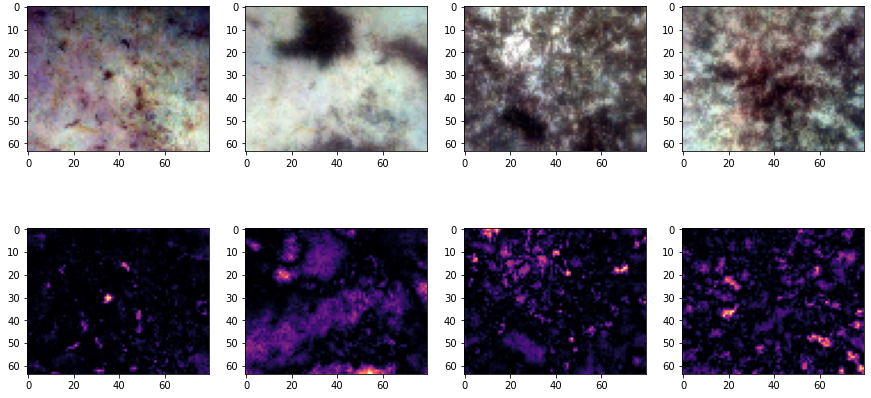}
         \caption{}
         \label{fig:inlier maps}
     \end{subfigure}
     \hfill
     \begin{subfigure}[b]{0.85\columnwidth}
         \centering
         \includegraphics[width=\textwidth]{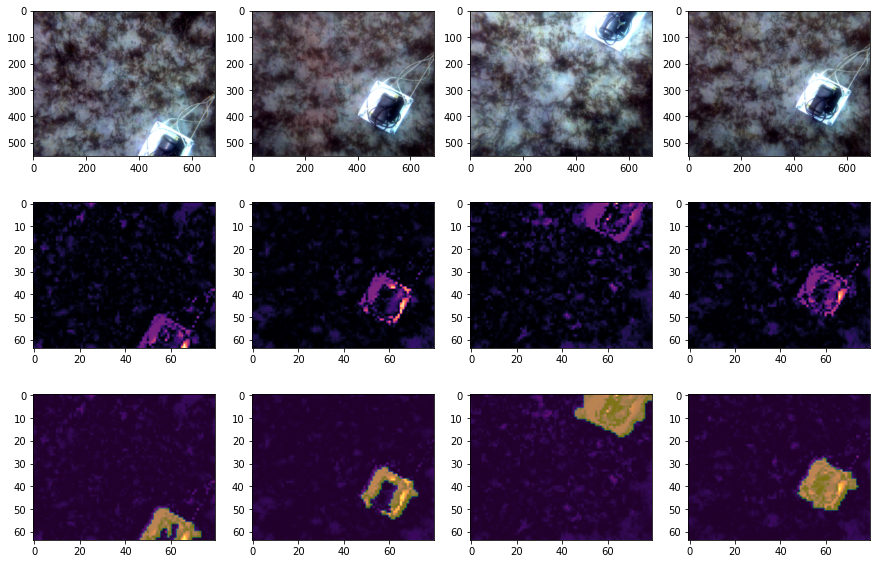}
         \caption{}
         \label{fig:outlier maps}
     \end{subfigure}
      \caption{(a) Inlier images and their corresponding anomaly maps; (b) Outlier images(top row), anomaly maps(middle row),ROIs extracted from anomaly maps(bottom row)}
        \label{fig:anomaly maps}
\end{figure}
Fig. \ref{fig:anomaly maps} shows the anomaly heatmaps of inlier images and outlier images. To automate the search for outliers, we therefore adopt an approach based on segmenting regions of each anomaly heatmap corresponding to high reconstruction error. 

\subsection{ROI extraction}

To generate the anomaly heatmap for each image, we compute the squared pixel-wise differences between the image and the corresponding reconstruction from the model. We extract a Region of Interest (ROI) for the anomaly heatmap in the following manner: A median blurring filter is applied to the heatmap before thresholding it to obtain a binary image. Erosion and dilation filters are to fill in small pits and other artefacts. This results in a binary image in which regions with high reconstruction error are prominently highlighted. Contours are extracted from the binary image and the area of each contour is computed. To exclude anomalous regions that are either too small or large relative to the object we are interested in detecting, lower and upper threshold values on the contour areas are applied. We apply these thresholds based on the expected size of the object to be detected. Given a known camera calibration and estimated vehicle altitude, the upper and lower size bounds of the object can be expressed in pixels. This additional contour-area criteria provides an additional means of discarding false-alarms. We use the term 'ROI anomaly score' to refer to the anomaly score for this method. For images in which the contour area criteria is not met, we assign an anomaly score of zero. When the criteria is met, we designate the mean value of reconstruction error within the contour boundary as the anomaly score. 
\begin{figure}[tb!]
  \includegraphics[width=0.85\linewidth]{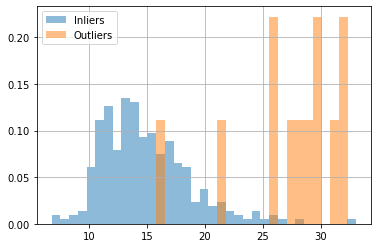}
  \caption{Distributions of ROI anomaly scores for the inlier and anomalous image classes}
  \label{fig: ROI score distributions}
\end{figure}
The distribution of ROI-based anomaly scores is shown in Fig. \ref{fig: ROI score distributions}. By applying this method to the test dataset, we are able to achieve a much larger degree of separation (relative to the previously shown L2 score) between the inliers and anomalous outliers. 

\subsection{Latent space clustering} \label{latent space clustering}

We analyze the latent spaces of the test set on our trained models. As we stated previously in \ref{detection metrics}, we have nine models, obtained by increasing the latent space dimensionality $d$ in 2x increments between $d$=8 and $d$=2048. \begin{figure}[tb!]
     \centering
     \begin{subfigure}[b]{0.80\columnwidth}
         \centering
         \includegraphics[width=\textwidth]{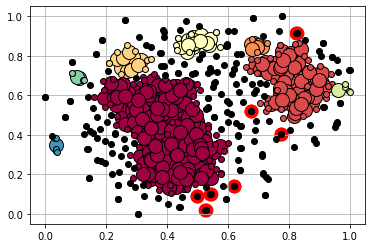}
         \caption{}
         \label{fig:dbscan}
     \end{subfigure}
     \hfill
     \begin{subfigure}[b]{0.80\columnwidth}
         \centering
         \includegraphics[width=\textwidth]{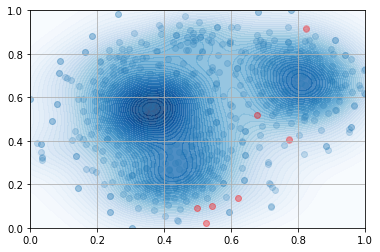}
         \caption{}
         \label{fig:kde}
     \end{subfigure}
     \hfill
     \begin{subfigure}[b]{0.80\columnwidth}
         \centering
         \includegraphics[width=\textwidth]{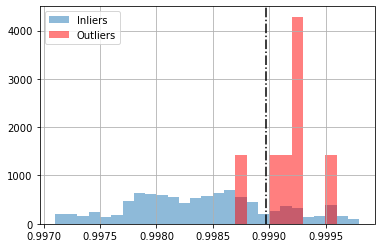}
         \caption{}
         \label{fig:kde score}
       \end{subfigure}  
      \caption{Latent space density estimation and clustering(a) Clustering using DBSCAN. The plot shows the obtained clusters and noise points(black). The anomalies are highlighted with a bold red outline; (b) A KDE fitted on the data. The anomalies are in red; (c) Histogram of densities from the fitted model. The chosen 80th-percentile threshold value is shown;}
        \label{fig:latent cluster}
\end{figure}
To reduce computational overhead due to the curse of dimensionality, and to simplify visualization, we reduce the dimensions to $d$ = 2. We choose t-Stochastic Neighbour Embedding (t-SNE) \cite{van2008visualizing} to arrive at a mapping from high-dimensional latent space to a 2D representation. Our anomaly detection method is based on the key observation that the anomalous outlier images tend to gather in the  low-density regions of the embedding. This lends itself well to density-based clustering algorithms. 
We evaluated the suitability of two algorithms to isolate the low density regions of the embedding. First, we apply Density-based spatial clustering of applications with noise (DBSCAN) \cite{ester1996density} to the test data. 
DBSCAN requires two parameters to be supplied by the user: Epsilon ($\epsilon$), the value of the radius that defines the neighborhood around each data point, and the minimum number of data points required in a neighborhood ($n_n$).  DBSCAN classifies data points into three categories: Core points, border points and noise points. Core points have at least $n_n$ data points within a neighborhood of radius $\epsilon$; Border points have less than $n_n$ neighbours,  but at least one core point in its neighborhood; Noise points are neither core nor border points. We use the following two well-established rules-of-thumb to assign values for $\epsilon$ and $n_n$: We assign $n_n$ as 2x$d_r$, where $d_r$ is the reduced number of dimensions of our dataset. To determine an optimal value of $\epsilon$, we find the so-called elbow of a k-distance plot as described in \cite{schubert2017dbscan}. 
Fig. \ref{fig:dbscan} shows the resulting clusters on the test data when we use a model with $d$=256. The algorithm assigns 425 datapoints to one of 8 clusters. The remaining 203 datapoints that are not members of any cluster are designated as noise points. As intended, we are able to significantly reduce the search-space of images for an operator to visually inspect. The outlier images tend to gather in the low density regions and can be isolated from the dense inlier regions. Although this approach yields positive results, we note a few shortcomings. We are required to tune two parameters, neither of which have any physically meaningful interpretation to the operator. Furthermore, points can be assigned to only two binary categories. We would instead prefer to obtain a probabilistic score in which the operator can assign a meaningful threshold. This would readily allow an outlier to be flagged based on statistical hypothesis tests. We therefore choose to apply Kernel Density Estimation (KDE) to fit a distribution to our dimensionally-reduced data.
For real values of our data $x$, the kernel density estimator is defined as \cite{silverman2018density}:
\begin{equation}
\widehat{f}_h(x)=\frac{1}{N h} \sum_{i=1}^N K\left(\frac{x-x_i}{h}\right)
\end{equation}
where $x_i$ are random samples from an unknown distribution. $N$ is the number of samples, $h$ is the bandwidth, and $K(.)$ is the kernel function, which we choose to be Gaussian:
\begin{equation}
K(u)=\frac{1}{\sqrt{2 \pi}} e^{\frac{-u^2}{2}}
\end{equation}
The choice of bandwidth is non-trivial and dictates the tradeoff between bias and variance of the fitted model. Rather than relying on heuristic methods as in \cite{silverman2018density}, we perform a grid search with 20-fold cross validation to select an appropriate bandwidth. Fig. \ref{fig:kde} shows the estimated density of the data. We evaluate each datapoint on the estimated PDF. We then choose an 80th-percentile threshold value on the density. As seen in Fig. \ref{fig:kde score}, we can separate out most of the anomalous images with the threshold. 

\section{RESULTS} \label{results}
\begin{figure}[tb!]
\begin{center}
  \includegraphics[width=0.80\linewidth]{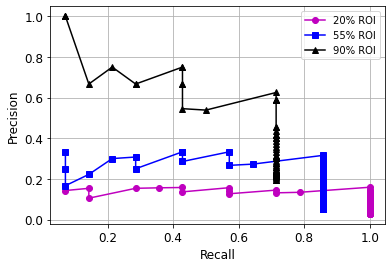}
\end{center}
  \caption{Precision-recall curves for $d$=2048. Each curve represents precision-recall tradeoff for a single $n^{th}$-percentile threshold on the ROI score}
  \label{fig: pr-curve}
\end{figure}

We first test the performance of our two-pronged approach in extracting a subspace of anomalous images, and discarding inlier images. We denote the inlier class as negative and the outlier class as positive. We use the following metrics to evaluate anomaly detection performance: precision, recall, and F1-score. Precision is defined as:
\begin{equation}
\textrm{Precision} = \frac{TP}{TP+FP}
\end{equation}
Recall is defined as:
\begin{equation}
\textrm{Recall} = \frac{TP}{TP+FN}
\end{equation}
The F1-score is defined as the harmonic mean of the precision and recall, weighting both measures equally. High precision relates to a low false positive rate, and high recall relates to a low false negative rate.
\begin{figure}[tb!]
     \centering
     \begin{subfigure}[htb!]{0.70\columnwidth}
         \centering
         \includegraphics[width=\textwidth]{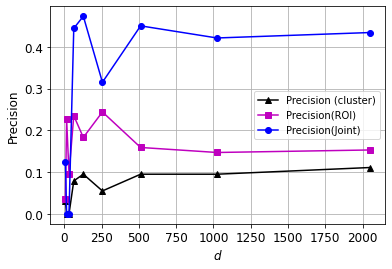}
         \caption{}
         \label{fig:precision results}
     \end{subfigure}
     \hfill
     \begin{subfigure}[htb!]{0.70\columnwidth}
         \centering
         \includegraphics[width=\textwidth]{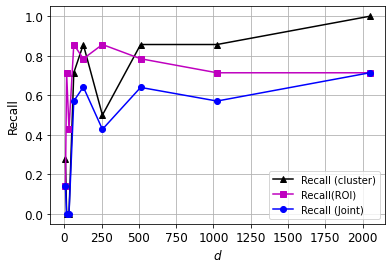}
         \caption{}
         \label{fig:recall results}
     \end{subfigure}
     \hfill
     \begin{subfigure}[htb!]{0.70\columnwidth}
         \centering
         \includegraphics[width=\textwidth]{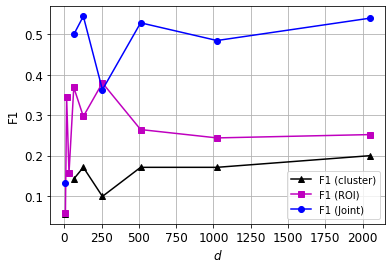}
         \caption{}
         \label{fig:F1 results}
     \end{subfigure}
      \caption{summary of results as a function of $d$ (a) Precision; (b) Recall; (c) F1 score. 80th percentile thresholds are applied on both the density scores and ROI scores}
        \label{fig:summary of results}
\end{figure}
We first evaluate the standalone performance of our clustering approach and ROI-based approach, before evaluating the joint performance of both methods.  We assess how sensitive the performance is to the latent dimensionality $d$.  Fig. \ref{fig:summary of results} shows the precision and recall for (1) our proposed clustering approach; (2) our proposed ROI-based method, and (3) both methods applied jointly, plotted as a function of latent dimensionality $d$. When we apply the clustering method on its own, we observe that precision is generally low across the models we tested. However, this is to be expected since the intention behind our clustering approach is to present a reduced subspace of images to the human operator. These images will then be subject to our ROI scoring method in the second stage to further prune away inlier image instances.
The fraction of relevant images presented to the human operator (recall) gradually increases with $d$. This occurs only beyond $d$=512. Below this value, both precision and recall fluctuate. We reason that models with low capacity are unable to capture visual features that would allow the outlier and inlier images to be separated in the reduced low-dimensional space. 
\begin{figure}[tb!]
\begin{center}
  \includegraphics[width=0.85\linewidth]{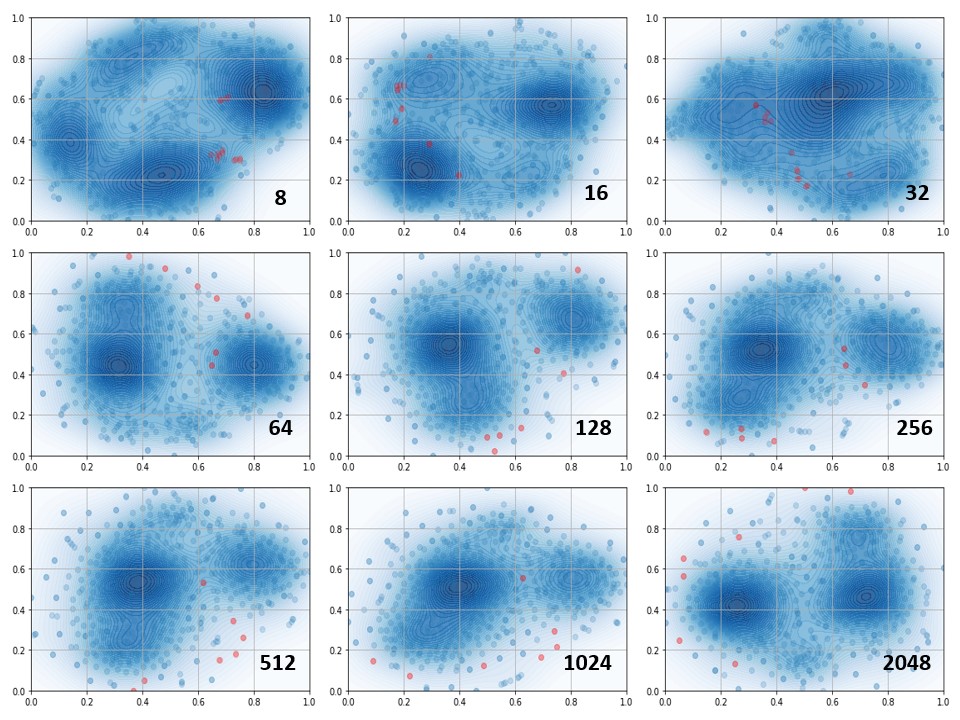}
\end{center}
  \caption{Latent space embeddings for models with varying latent vector dimensions. The dimensionality is shown in each subplot.}
  \label{fig: kde cluster summary}
\end{figure}
This is supported by Fig. \ref{fig: kde cluster summary}, in which we plot the reduced latent space for each model. Low values of $d$ result in the outliers being clustered together with inliers. 
Precision and recall for our ROI-based approach applied on its own  is not as sensitive to the size of the latent dimension $d$, and plateaus beyond $d$=512.  The recall is generally lower than the clustering method but this is at the expense of far greater precision. To apply both methods jointly, we first apply the clustering method to obtain a reduced subspace and then apply the ROI scores to the reduced subspace. We are able to obtain a much lower false alarm rate (precision=0.434), with recall=0.714. To visualize the tradeoff between precision and recall and dependency on ROI thresholds, the precision-recall curve for a model with $d$=2048 is shown in Fig. \ref{fig: pr-curve}. The figure also shows the effect of varying the ROI threshold. A very low threshold for the ROI score results in a near-stagnant value of precision. Increasing the ROI threshold shifts the PR-curve upwards and towards the left, essentially trading off recall for better precision. We obtain the largest average Precision (approximate measure of area under PR-curve) of 0.64 for a 95th-percentile threshold on the ROI score.

\section{CONCLUSIONS}

We present an anomaly detection system based on VAEs which can be used to detect artificial objects in large datasets of seafloor imagery. Estimating the density of the latent space provides a convenient method for a human operator to systematically query a database by simply applying an nth-percentile threshold. False positives within the returned images can be further filtered out by applying our ROI-based method. We test the performance of our approach as a function of the latent dimensionality of the VAE, and find that the clustering approach is particularly sensitive to the number of latent dimensions $d$. As future work, we plan to apply a transfer-learning approach to our problem by using pre-trained models as feature extractors in the VAE encoder. Furthermore, in this paper we did not consider temporal dependencies between images or their latent representations. Extending our approach to image sequences would potentially support online anomaly detection. Another interesting area of future work would be in adapting reconstruction error thresholds rather than maintaining a static threshold across different datasets. This would be a first step towards developing an adaptive system that is robust to changes in the environment or operating conditions.

\section{ACKNOWLEDGEMENTS}
The research for this paper received funding from the Australian Government through Trusted Autonomous Systems, a Defence Cooperative Research Centre funded through the Next Generation Technologies Fund. The authors thank the captain and crew of the Research Vessel \textit{Kimbla} for supporting the AUV missions and experimental campaign at Jervis Bay. The authors also thank Dr. Phillip Skelton for his feedback on the manuscript.
\addtolength{\textheight}{-12cm}   




\printbibliography 

@string{tim = "{IEEE Transactions on Instrumentation and Measurement}"}

@string{iros  = "IEEE/RSJ Intl. Conf. on Intelligent Robots and Systems (IROS)"}

@string{ieee  = "Proc. of the IEEE"}

@article{teng2020underwater,
  title={Underwater target recognition methods based on the framework of deep learning: A survey},
  author={Teng, Bowen and Zhao, Hongjian},
  journal={Intl. J. of Advanced Robotic Systems},
  volume={17},
  number={6},
  pages={1729881420976307},
  year={2020},
  publisher={SAGE Publications Sage UK: London, England}
}

@inproceedings{yu2018man,
  title={Man-made object recognition from underwater optical images using deep learning and transfer learning},
  author={Yu, Xian and Xing, Xiangrui and Zheng, Han and Fu, Xueyang and Huang, Yue and Ding, Xinghao},
  booktitle={2018 IEEE Intl. Conf. on Acoustics, Speech and Signal Processing (ICASSP)},
  pages={1852--1856},
  year={2018},
  organization={IEEE}
}

@inproceedings{abati2019latent,
  title={Latent space autoregression for novelty detection},
  author={Abati, Davide and Porrello, Angelo and Calderara, Simone and Cucchiara, Rita},
  booktitle={Proc. of the IEEE/CVF Conf. on Computer Vision and Pattern Recognition},
  pages={481--490},
  year={2019}
}

@article{fei2020attribute,
  title={Attribute restoration framework for anomaly detection},
  author={Fei, Ye and Huang, Chaoqin and Jinkun, Cao and Li, Maosen and Zhang, Ya and Lu, Cewu},
  journal={IEEE Trans. on Multimedia},
  year={2020},
  publisher={IEEE}
}

@inproceedings{zong2018deep,
  title={Deep autoencoding gaussian mixture model for unsupervised anomaly detection},
  author={Zong, Bo and Song, Qi and Min, Martin Renqiang and Cheng, Wei and Lumezanu, Cristian and Cho, Daeki and Chen, Haifeng},
  booktitle={Intl. conf. on learning representations},
  year={2018}
}

@article{krizhevsky2012imagenet,
  title={Imagenet classification with deep convolutional neural networks},
  author={Krizhevsky, Alex and Sutskever, Ilya and Hinton, Geoffrey E},
  journal={Adv. in neural information processing systems},
  volume={25},
  year={2012}
}

@article{rao2017multimodal,
  title={Multimodal learning and inference from visual and remotely sensed data},
  author={Rao, Dushyant and De Deuge, Mark and Nourani--Vatani, Navid and Williams, Stefan B and Pizarro, Oscar},
  journal={The Intl. J. of Robotics Research},
  volume={36},
  number={1},
  pages={24--43},
  year={2017},
  publisher={SAGE Publications Sage UK: London, England}
}

@inproceedings{flaspohler2017feature,
  title={Feature discovery and visualization of robot mission data using convolutional autoencoders and Bayesian nonparametric topic models},
  author={Flaspohler, Genevieve and Roy, Nicholas and Girdhar, Yogesh},
  booktitle={2017 IEEE/RSJ Intl. Conf. on Intelligent Robots and Systems (IROS)},
  pages={1--8},
  year={2017},
  organization={IEEE}
}

@article{yamada2021learning,
  title={Learning features from georeferenced seafloor imagery with location guided autoencoders},
  author={Yamada, Takaki and Pr{\"u}gel-Bennett, Adam and Thornton, Blair},
  journal={J. of Field Robotics},
  volume={38},
  number={1},
  pages={52--67},
  year={2021},
  publisher={Wiley Online Library}
}

@article{zurowietz2018maia,
  title={MAIA—A machine learning assisted image annotation method for environmental monitoring and exploration},
  author={Zurowietz, Martin and Langenk{\"a}mper, Daniel and Hosking, Brett and Ruhl, Henry A and Nattkemper, Tim W},
  journal={PloS one},
  volume={13},
  number={11},
  pages={e0207498},
  year={2018},
  publisher={Public Library of Science San Francisco, CA USA}
}

@article{huang2021self,
  title={Self-supervision-augmented deep autoencoder for unsupervised visual anomaly detection},
  author={Huang, Chao and Yang, Zehua and Wen, Jie and Xu, Yong and Jiang, Qiuping and Yang, Jian and Wang, Yaowei},
  journal={IEEE Trans. on Cybernetics},
  year={2021},
  publisher={IEEE}
}

@article{bozcan2021gridnet,
  title={Gridnet: Image-agnostic conditional anomaly detection for indoor surveillance},
  author={Bozcan, Ilker and Le Fevre, Jonas and Pham, Huy X and Kayacan, Erdal},
  journal={IEEE Robotics and Automation Letters},
  volume={6},
  number={2},
  pages={1638--1645},
  year={2021},
  publisher={IEEE}
}

@article{mantegazza2022outlier,
  title={An Outlier Exposure Approach to Improve Visual Anomaly Detection Performance for Mobile Robots},
  author={Mantegazza, Dario and Giusti, Alessandro and Gambardella, Luca Maria and Guzzi, J{\'e}r{\^o}me},
  journal={IEEE Robotics and Automation Letters},
  volume={7},
  number={4},
  pages={11354--11361},
  year={2022},
  publisher={IEEE}
}

@article{chen2020unsupervised,
  title={Unsupervised lesion detection via image restoration with a normative prior},
  author={Chen, Xiaoran and You, Suhang and Tezcan, Kerem Can and Konukoglu, Ender},
  journal={Medical image analysis},
  volume={64},
  pages={101713},
  year={2020},
  publisher={Elsevier}
}

@inproceedings{havtorn2021hierarchical,
  title={Hierarchical vaes know what they don’t know},
  author={Havtorn, Jakob D and Frellsen, Jes and Hauberg, S{\o}ren and Maal{\o}e, Lars},
  booktitle={Intl. Conf. on Machine Learning},
  pages={4117--4128},
  year={2021},
  organization={PMLR}
}

@article{kingma2013auto,
  title={Auto-encoding variational bayes},
  author={Kingma, Diederik P and Welling, Max},
  journal={arXiv preprint arXiv:1312.6114},
  year={2013}
}

@article{van2008visualizing,
  title={Visualizing data using t-SNE.},
  author={Van der Maaten, Laurens and Hinton, Geoffrey},
  journal={J. of machine learning research},
  volume={9},
  number={11},
  year={2008}
}

@inproceedings{ester1996density,
  title={A density-based algorithm for discovering clusters in large spatial databases with noise.},
  author={Ester, Martin and Kriegel, Hans-Peter and Sander, J{\"o}rg and Xu, Xiaowei and others},
  booktitle={kdd},
  volume={96},
  number={34},
  pages={226--231},
  year={1996}
}

@book{silverman2018density,
  title={Density estimation for statistics and data analysis},
  author={Silverman, Bernard W},
  year={2018},
  publisher={Routledge}
}

@article{schubert2017dbscan,
  title={DBSCAN revisited, revisited: why and how you should (still) use DBSCAN},
  author={Schubert, Erich and Sander, J{\"o}rg and Ester, Martin and Kriegel, Hans Peter and Xu, Xiaowei},
  journal={ACM Trans. on Database Systems (TODS)},
  volume={42},
  number={3},
  pages={1--21},
  year={2017},
  publisher={ACM New York, NY, USA}
}

@article{scott1991feasibility,
  title={Feasibility of multivariate density estimates},
  author={Scott, David W},
  journal={Biometrika},
  volume={78},
  number={1},
  pages={197--205},
  year={1991},
  publisher={Oxford University Press}
}

\end{document}